\documentclass[11pt,a4paper]{article}
\usepackage[hyperref]{eacl2021}
\usepackage{times}
\usepackage{latexsym}

\usepackage{booktabs}
\usepackage{graphicx}
\usepackage{amsmath}

\usepackage{microtype}

\aclfinalcopy % Uncomment this line for the final submission
\newcommand{\coqa}{CoQA}

\newcommand{\name}{\textit{ChainCQG}}

\title{ChainCQG: Flow-Aware Conversational Question Generation}
\author{
Jing Gu\textsuperscript{\rm 1,3} \,\,\, Mostafa Mirshekari\textsuperscript{\rm 1} \,\,\, Zhou Yu\textsuperscript{\rm 1,2} \,\,\, Aaron Sisto\textsuperscript{\rm 1} \\
\textsuperscript{\rm 1} Searchable.ai \\
\textsuperscript{\rm 2} Columbia University \\
\textsuperscript{\rm 3} University of California, Davis \\
\{jing,mostafa,aaron\}@searchable.ai \\
zy2461@columbia.edu
}

\date{}

\begin{document}
\maketitle

\begin{abstract}

Conversational systems enable numerous valuable applications, and question-answering is an important component underlying many of these. However, conversational question-answering remains challenging due to the lack of realistic, domain-specific training data. Inspired by this bottleneck, we focus on conversational question generation as a means to generate synthetic conversations for training and evaluation purposes. 
% Here, we present \name{}, a neural Conversational Question Generation~(CGQ) model and evaluate its performance on both automatic metrics and human metrics.
We present a number of novel strategies to improve conversational flow and accommodate varying question types and overall fluidity. Specifically, we design \name{} as a two-stage architecture that learns question-answer representations across multiple dialogue turns using a flow propagation training strategy. \name{} significantly outperforms both answer-aware and answer-unaware SOTA baselines~(e.g., up to 48\% BLEU-1 improvement). Additionally, our model is able to generate different types of questions, with improved fluidity and coreference alignment.
\end{abstract}

\section{Introduction}\label{sec:intro}

Conversational systems are important in many real-world applications, including personal assistants, educational tutors~\citep{winkler2020sara}, customer service~\citep{asri2017frames, budzianowski2018multiwoz}, and increasingly, entertainment. A key component of these systems is the ability to interpret a search query and retrieve information from different sources as naturally and efficiently as possible. In analogous human interactions, such a search generally occurs through conversation. In this context, a conversation consists of a sequence of dialogue turns during which the search objective becomes clearer over time. The applications mentioned above could benefit greatly from this type of multi-turn interaction, enabling conversational agents to accurately predict intent, request additional information, and better understand ambiguous followup questions and comments. In an applied setting, meaningful and natural conversations are important features of virtual entities as a means to establish trust and improve usability.

Here, we are motivated by the challenging task of conversational question answering~(CQA). Current open-source datasets such as CoQA~\citep{reddy2019coqa} and QuAC~\citep{choi2018quac} provide strong baselines for this task. However, these datasets have limited applicability in practical settings, because 1)~they are created from domain-agnostic source material, and 2)~they do not necessarily consider the full diversity of question types and vernacular that may be encountered in natural dialogue. Creating realistic, domain-specific datasets to train CQA models is notoriously costly and time-consuming. As such, we focus on the related task of question generation as a means to generate synthetic conversational questions  and subsequently, create new datasets or augment existing ones. This will ultimately enable training CQA models in closed-loop, simulation environments, as well as allow machines to initiate dialogue and engage in information-seeking behavior.

While QA models have been studied previously~\citep{zhu2018sdnet, huang2018flowqa, yeh2019flowdelta, chen2019graphflow, ju2019technical, ohsugi2019simple}), the QG task, which is the focus of this paper, has received less attention. QG models in the answer-unaware setting aim to predict a question given the source passage, while in the answer-aware setting, the target answer and rationale are included as inputs as well. Most QG-related literature has focused on single-turn question generation using question-answer datasets such as SQuAD ~\citep{rajpurkar2016squad}, and other textual sources like Wikipedia articles ~\citep{du2018harvesting}. 

Conversational Question Generation~(CQG) proves more challenging than single-turn QG as the questions are often highly ambiguous on their own, forcing the model to learn a deeper understanding of the context surrounding the passage text and dialogue history ~\citep{pan2019reinforced}. Most CQG studies have generated questions using only the passage and dialogue history as inputs~(i.e., answer-unaware)~\citep{pan2019reinforced, qi2020stay, nakanishi2019towards, wang2018learning}. Answer-aware CQG models, on the other hand, generate questions based on the target answer, as well as dialogue history and passage. Although answer-aware CQG models seek to improve the generated conversation flow, current answer-aware QG models suffer from issues including inaccurate coreference alignment, dialogue inconsistencies, incorrect grammar and the inability to generate many different types of questions (e.g. yes/no, factoid, explanation).

% What about just using the answer concatendated with context.
In this paper, we introduce \name{}, a Conversational QG model that achieves improved performance by jointly learning the representations of questions and answers sequentially, across multiple dialogue turns. To this end, we outline a two-stage architecture, inspired by the approach discussed in~\citet{wu2019alternating} and \citet{gu2020tailored}, where two language models are used to simulate user and system in a response generation task.
%each module is a GPT-2 model~\citep{radford2019language}
Our \name{} model is trained end-to-end, resulting in high-quality questions, while reducing computational cost by using shared parameters across both models. Using an answer-aware strategy grounds each turn of QG by jointly encoding the passage with the target answer rationale, increasing accuracy of the generated question types and further aligning coreferences between dialogue turns. We evaluate our approach using the inverted CoQA dataset~\citep{reddy2019coqa}, which is a large-scale CQA dataset that we re-purposed for question generation. Our model outperforms existing SOTA CFNet~\citep{gao2019interconnected} and ReDR~\citep{pan2019reinforced} by a large margin on automatic evaluation metrics, and shows improved results on human evaluation metrics as well. More information about the baselines will be discussed in Section~\ref{sec:related}.

In summary, the main contributions of this paper are threefold \footnote{Code available at \url{https://github.com/searchableai/ChainCQG.}}:
\begin{itemize}
\item The \name{} two-stage architecture is introduced with answer-aware input encoding, and it is an end-to-end model which is able to fluently generate different types of questions and achieve high consistency with the target answers.

\item We demonstrate a flow propagation-based training method to learn question-answer representations across multiple dialogue spans.
\item \name{} sets the new SOTA results on the answer-aware CQG task with robust human evaluation results.
\end{itemize}

 The remainder of the paper is structured as follows: First, we discuss related work and how our approach is distinguished from previous methods~(in Section~\ref{sec:related}). Then, we discuss the \name{} framework and preprocessing steps~(in Section~\ref{label:method}). Next, we describe our experiments, datasets and metrics, and evaluation results~(in Section~\ref{sec:experiments}). Finally, we discuss conclusions, future work and the ethical issues~(in Sections~\ref{sec:ethical} and~\ref{sec:conclusion}).

\section{Related Work}\label{sec:related}
In this section, we first explore previous approaches to question generation and then discuss outstanding research challenges in conversational question generation.

\subsection{Single-turn Question Generation}

Single-turn question generation has been the focus of extensive research. Two of the main categories in QG are answer-unaware and answer-aware. The former category generates the question without knowledge of the answer and solely based on the passage; whereas, the latter takes both passage and answer as inputs. Traditional approaches for answer-unaware QG include two main steps: content selection and question generation~\citep{du2017identifying, subramanian2017neural}. Some of the more recent approaches utilize sequence-to-sequence~(seq2seq) models for end-to-end question generation using Transformer-based architectures~\citep{scialom2019self}. Various techniques have been used for improving the generated questions, including contextualized word embeddings~\citep{scialom2019self}, question type usage and copying mechanism~\citep{wu2020question}, and typed decoders~\citep{wang2018learning}.

To enable answer-aware question generation, the input passage is augmented with information describing the answer. For example, the passage can be concatenated with the answer positions and lexical features~(e.g., part-of-speech (POS) and named entity (NER)) to form the encoder input of a seq2seq model~\citep{zhou2017neural}. Jointly modelling the unstructured passage and the structured answer-relevant relation has been suggested for improving question generation as well~\citep{li-etal-2019-improving-question}. Additional techniques have been proposed to solve various answer-aware QG challenges, including poor performance on long passages~\citep{zhao2018paragraph} and the bias of repeating the terms in the target answer within the generated question~\citep{kim2019improving}. 

\subsection{Conversational Question Generation}

Compared to single-turn QG, conversational~(i.e., multi-turn) QG is less frequently explored in the literature. Further, it is more difficult as it requires a deeper understanding of the context and the dialogue history. Previous work mostly focused on answer-unaware CQG~\citep{pan2019reinforced, qi2020stay, nakanishi2019towards, wang2018learning}. Specifically, ~\citet{pan2019reinforced} proposed an encoder-decoder framework, ReDR, for answer-agnostic CQG, which is fine-tuned using feedback from an independent question-answer model. However, in this setting, maintaining conversational flow and consistency between dialogue turns is a primary challenge. 

In this paper, we focus on answer-aware question generation. By grounding the generated question with the target answer rationale in each turn, this approach seeks to improve conversational flow and question-answer consistency. Within answer-aware CQG, ~\citep{gao2019interconnected} introduced the current SOTA, CFNet, which combined an auxiliary coreference alignment module with a copy mechanism and dialogue flow embedding. As will be described ~(in Section~\ref{sec:experiments}), we compare our model to answer-aware, CFNet~\citep{gao2019interconnected}, and answer-unaware, ReDR~\citep{pan2019reinforced}, SOTA baselines. As a practical note, in real-world applications, answer-aware QG systems may be augmented with an Answer Generation~(AG) model to form an answer-unaware model that predicts the next conversational question-answer pair jointly. Discussing this AG model is out of the scope of this paper and will be the focus of future work.

\section{The \name{} ~Framework}\label{label:method}

\name{} learns the question-answer representations jointly using two modules: Answer Encoding~(AE) and Question Generation~(QG). Encoding the answer based on the passage and dialogue history improves the answer understanding within the QG module, which in turn improves the generated questions. In the rest of this section, we provide a description of the input pre-processing steps, the general CQG problem formulation, and the AE and QG modules.

\begin{figure*}
    \centering
    \includegraphics[scale=0.25]{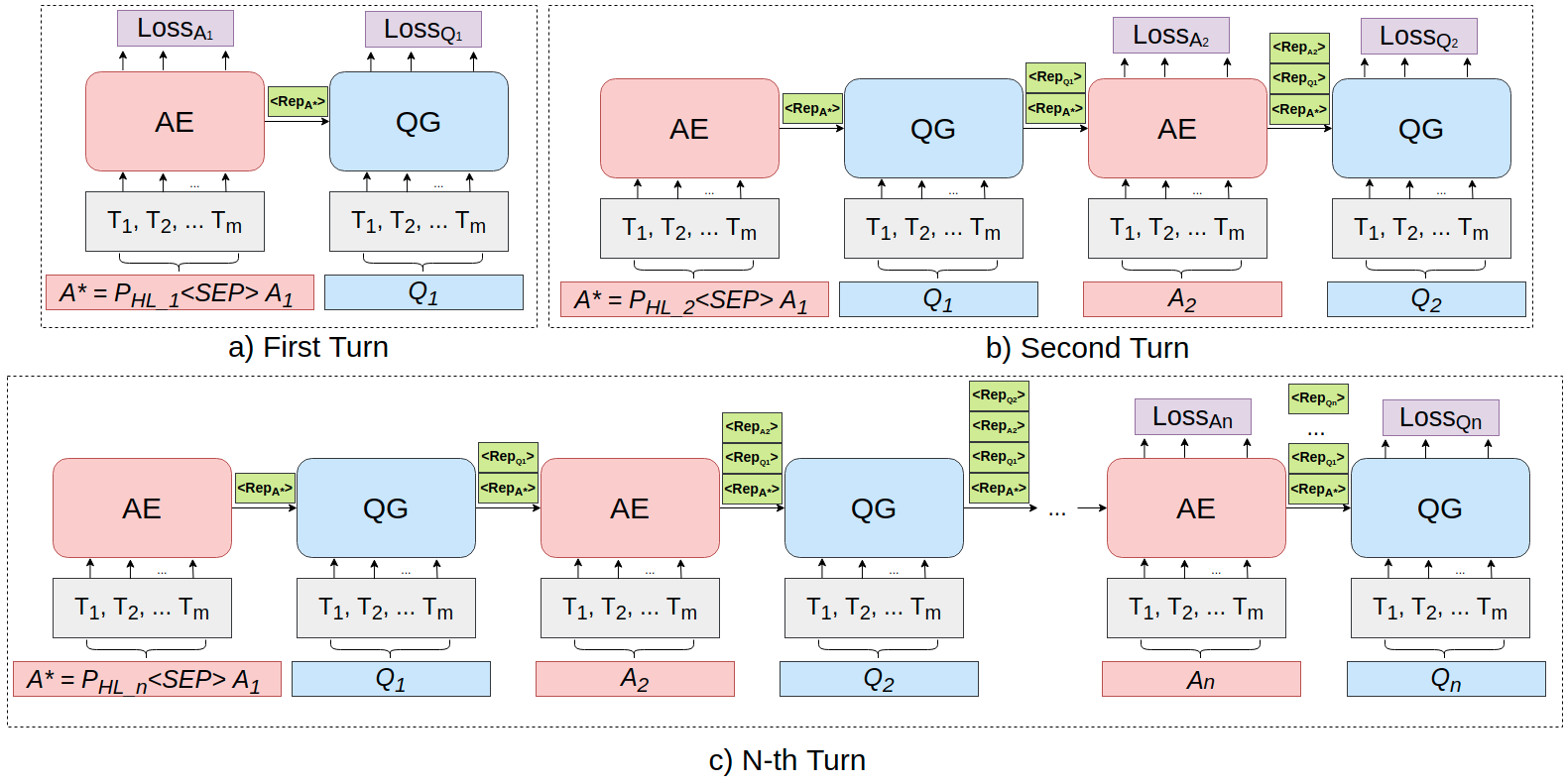}
    \caption{Main structure of \name{}. Each QA turn in the dialogue span is trained with a separate conversational flow that contains all previous dialogue turns. Answer Encoder and Question Generator modules iteratively generate and share answer and question representations across multiple dialogue turns.}
    \label{fig:model}
\end{figure*}

\subsection{Task Definition}

The conversational QG task in this paper aims to predict the next question given the passage (P), target answer ($A_n$) and history of the dialogue preceding the n\textsuperscript{th} turn, ($H_n$). We also consider the answer rationale in each turn, and annotate the passage with the target answer rationale span, which we denote as $P_{HL_n}$. Mathematically, given the annotated passage~($P_{HL_n}$), target answer~($A_n$), and dialogue history~($H_n=({(Q_{1}, A_{1}), (Q_{2}, A_{2}), ..., (Q_{n-1}, A_{n-1})})$), the QG task predicts the next question $Q_n$. This task can be defined explicitly as generating a question, $\hat{Q}$, where:
\begin{equation}
    \hat{Q} = \underset{Q_n}{argmax} Prob(Q_n|P_{HL_n}, A_n, H_n).
\end{equation}

\subsection{Input Preprocessing}\label{sec:preprocess}

In this Section, we briefly describe the processing steps necessary to prepare the input data. Specifically, we take the following approach:

\begin{enumerate}
    \item We first create $n$ sub-dialogues based on the full dialogue, with the $i$-th sub-dialogue starting from the first turn and finishing with the $i$-th turn, i.e., $SD_1$ = \{\{$Q_1$, $A_1$\}\}, $SD_2$ = \{\{$Q_1$, $A_1$\}, \{$Q_2$, $A_2$\}\}, ..., $SD_n$ = \{\{$Q_1$, $A_1$\}, \{$Q_2$, $A_2$\}, ..., \{$Q_n$, $A_n$\} \}.
    \item For the $i$-th sub-dialogue, we use a highlight token, [HL], to denote the answer rationale in the passage corresponding to the answer in the $i$-th turn, which serves as additional context for the target answer. 
    \item For each sub-dialogue, $i$, the passage (with the highlighted token corresponding to the $i$-th turn) is concatenated with the first answer (i.e., $A_1$) by a SEP token. We denote the concatenation as $A^*$.
	\item We then reverse the order of the answers and questions in the sub-dialogues~(e.g., \{A1, Q1\} instead of \{Q1, A1\}). The reason behind this step is to align the input sequence with the natural order of the QG task, where the questions come after the answers. We examine the effects of this ordering scheme in later ablation studies.
\end{enumerate}

\subsection{Answer Encoding and Question Generation Modules}
\label{sec:AEQG}

In this paper, we use GPT-2 \citep{radford2019language} to represent both the AE and QG modules described above. Specifically, the AE model is used to learn the representation of the passage and answer in each turn, and the QG model is used to learn the representation of the dialogue history and generate the next question in the conversation. The AE and QG models communicate via the models' hidden states, which are the K and V values when using GPT. K and V values together form a contexual representation for the entire conversation history. 
%We use cross entropy loss to maximize the probability of the target question, .

\subsection{Flow Propagation-Based Training} 
To improve the conversational flow of the CQG, we introduce a sequential training process. Figure~\ref{fig:model} shows the main structure. To make it more concrete, let us consider a dialogue of $n$ turns~(\{\{$Q_1$, $A_1$\}, \{$Q_2$, $A_2$\}, ..., \{$Q_n$, $A_n$\} \}).
The forward propagation process iterates through all previous turns and finally estimates the loss values for $A_n$ and $Q_n$. In this process, we pass the GPT-based (K, V) representation forward to the next module, which accumulates the representations of each previous turn, with the original highlighted passage as reference.
For each sub-dialogue, we only update the loss from the answer and question in the last turn since the highlighted span specifies the information for the last turn. For a sub-dialog of $n$ turns, the loss is calculated as 

\begin{equation}
Loss = Loss_{A_n} + Loss_{Q_n}
\end{equation}
where
\begin{equation}
    Loss_{A_n} = CE(A_n, P_{A_n})
\end{equation}

and 
\begin{equation}
    Loss_{Q_n} = CE(Q_n, P_{Q_n})
\end{equation}
CE refers to the cross-entropy loss from a target sentence.
The parameters of the model are updated by backpropagating the aggregated loss values. By considering the encoding of the previous turns for estimating the loss and increasingly considering various sub-dialogues, the flow propagation-based training improves the conversational flow of the CQG.

\section{Experiments}\label{sec:experiments}

\subsection{Dataset}

We conduct experiments on the \coqa{} dataset \citep{reddy2019coqa}, which is a large-scale conversational question answering dataset composed of 8k conversations with 127k question-answer pairs collected via Amazon Mechanical Turk. Each dialogue turn also contains the supporting rationale for each answer. A number of different question types are present, as documented in the original paper, including Yes/No, explanation (i.e. How?, Why?), and factoid (i.e. What? When? Where? How much?). Yes/No and explanation questions, alongside the overall conversational language, make this an exceedingly challenging dataset for CQG.
Since the private test set is not available, we conduct experiments on the training set and the dev set. We randomly sample 10\% from the original training set to form the test set, and keep the original dev set unchanged. We conduct our experiments with a training set with 97783 examples, dev set with 7983 examples and test set with 10846 examples. We report the performance on the test set. 

\subsection{Implementation Details}

We use both GPT\textsubscript{small} and GPT\textsubscript{medium} in all experiments. For baselines, we consider ReDR, the SOTA method in answer-unaware CQG, and CFNet, the SOTA method in answer-aware CQG. We also implemented two SOTA pre-training generation models, T5 and BART. They utilize all our preprocessing methods and training skills except the AE/QG modules. We used T5\textsubscript{large} (770M) and BART\textsubscript{large} (400M), which are comparable with \name{}-M in terms of parameter size. \par

We initialize \name{} with the open sourced GPT-2 parameters~\citep{radford2019language}. We apply AdamW optimizer~\citep{adamw}, and the warmup ratio is set to 0.1. The learning rate is tuned between 2e-5 and 5e-5. The dropout ratio is set to be 0.1. We decode questions by nucleus sampling~\citep{nucleus_sampling} with top-p as 0.2, top-k as 400, and temperature as 0.7.

 %\citep{gao-etal-2019-interconnected}

\subsection{Evaluation Metric}\label{sec:evalmetrics}

Our main objectives when evaluating our model are quality of the generated questions and performance on our task goal~(e.g., asking conversational questions that are consistent with the target answers). To this end, we first examine a set of automated metrics. Then, to ensure robustness, we evaluate and discuss a set of human-based metrics.

\subsubsection{Automated Metrics}

To evaluate our question generation approach, we aim to show that it is 1)~grammatically and semantically correct and 2)~able to achieve the task objectives. To achieve the first goal, we compute automatic metrics with respect to the ground truth questions. We report multiple commonly used metrics, including BLEU~\citep{papineni2002bleu}, ROUGE~\citep{lin2004rouge}, METEOR~\citep{banerjee2005meteor}, and perplexity~\citep{clarkson1999towards}.
BertScore~\citep{zhang2019bertscore} and MoverScore~\citep{zhao2019moverscore} are recently proposed metrics that utilize a large pre-trained model to evaluate the generation quality at the semantic level. We use both of these to evaluate the semantic similarity between the generated question and the reference question. Since two questions could express similar meaning with low lexical overlap, these two semantic-level metric could also show important information about the question quality.

\subsubsection{Human Evaluation Metrics and Procedure}

In this section, we discuss the metrics used for human evaluation. Human evaluation provides additional support for the approach and the robustness of automatic evaluation. Specifically, we use answerability and fluency to measure the quality of the generated questions in relation to the context. We have used Mechanical Turk for this evaluation.

In the context of answer-aware question generation, \textbf{Answer Consistency} describes whether the generated questions result in the correct answers~\citep{evaluation_survey}. To measure this metric, we provide the passage and the answer, and ask the evaluators whether the generated question is consistent with the answer~(i.e., $1$ for consistent and $0$ for inconsistent).
\textbf{Fluency} measures the quality of the generated questions and accounts for criteria such as grammar, spelling, choice of words, and style~\citep{du2017learning}. To measure this metric, we provide the generated question and ask the human evaluator whether the language in the generated question is fluent. We consider three categories of errors: grammar/spelling mistakes, missing entity names, and mismatched pronouns. Based on these categories, we assign $2$ for cases with no mistake in any of the categories, $1$ for cases with maximum of one mistake in any of the mentioned categories, and $0$ for cases with one or more mistakes in each one of the categories. We scale the fluency score to (0,1) by maximum evaluation scores.

\subsection{Main Results}

%1. Main table, introducing our performance with the baseline methods.

\begin{table*}
    \centering
     \begin{tabular}{c c c c c c c c c} 
     \hline
      Model & B1& B2& B3& B4 & M & RL & BS & MS \\
     \hline
      ReDR & 27.58 & 7.81 & 2.83 & 1.35 & 12.15 & 34.05 & 87.14 & 7.62\\ 
     \hline
     CFNet & 38.24 & 22.60 & 16.11 & 12.23 & 25.75 & 43.25 & 91.25 & 25.92 \\
     \hline
     BART-large & 49.41 & 30.57 & 19.40 & 12.34 & 35.78 & 46.88 & 92.55 & 31.89 \\ 
     \hline
      T5-large & 50.83 & 32.64 & 20.81 & 13.84 & 37.08 & 48.67 & 92.86 & 33.91\\
     \hline
      \name{}-M & \textbf{53.15} & \textbf{35.31} & \textbf{23.31} & \textbf{15.78} &  \textbf{40.15} & \textbf{50.98} & \textbf{93.14} & \textbf{36.40} \\
      \name{}-S & 49.26 & 31.06 & 20.24 & 12.11 & 33.26 & 46.23 & 92.53 & 32.82 \\
     \hline
    \end{tabular}
    \caption{Automated Metric Evaluation Results.}
    \label{tab:main_results}
\end{table*}

The \name{} model architecture is evaluated alongside two SOTA baselines (ReDR, CFNet) on a number of automatic metrics, including BLEU (1-4), METEOR, ROUGE-L, MoverScore, and BERTScore. More information about these metrics and baselines was presented in Section~\ref{sec:evalmetrics}. These scores seek to evaluate the lexical overlap, and to some degree, the semantic similarity, between generated and ground truth questions within each dialogue turn. We also train two QG models based on pretrained Transformer seq2seq architectures (BART-large~\citep{lewis-etal-2020-bart}, T5-large~\cite{T5}) using all elements of the \name{} methodology except the question-answer representation sharing mechanism used in \name{} model. Instead, the target answer is the direct input to the model, concatenated with the passage and dialogue history.

\subsubsection{Automated Metrics Results}
Results of all models and baselines are shown in Table~\ref{tab:main_results}. In the top row of this Table, P is the perplexity, B1-4 are BLEU 1 through BLEU 4, M is METEOR, RL is ROUGE-L, BS is the BERTScore, and MS is MoverScore. In the first column, ChainCQG-M and ChainCQG-S refer to two version of our approach using medium and small GPT-2. Major observations are listed below:

The top performing \name{} model, composed of two GPT-2 Medium modules, improves upon all baselines by a considerable margin and across all the considered metrics. In addition, it also outperforms the T5-large, which has more parameters, by a large margin. This suggests that the \name{} learns a better representation using the AE-QG structure, with less parameters than the T5.
We improve upon the current answer-aware CQG SOTA, CFNet, on each metric as well. Note that with our methods, T5-large and BART-large also outperforms the SOTA methods.
T5-large is the next best performing model, trailing \name{} by at least two points on all metrics except BERTScore, which shows a narrower margin of improvement.

\subsubsection{Human Evaluation Results}
\begin{table}
    \centering
     \begin{tabular}{c c c} 
     \hline
     Model & Consistency & Fluency \\
     \hline
     CFNet  & 0.710 & 0.439 \\
     \hline
     BART   & 0.792 & 0.482 \\
     \hline
     T5  & 0.757 & 0.462 \\
     \hline
     \name{}-M & \textbf{0.817} & \textbf{0.548} \\
     \hline
    \end{tabular}
    \caption{Human Evaluation Results.}
    \label{tab:human_eval}
\end{table}

It is well-known that automatic evaluation metrics do not always correlate with human judgement in conversational generation tasks \citep{evaluation_survey}. Especially in the context of CQG, there is a many-to-one relationship between questions and their target answers and dialogue contexts, and token-based metrics are inherently unable to measure the similarity between such sequences with low degrees of lexical overlap. As a recourse, we also assess our model performance on a number of human evaluation metrics described in a previous section: Answer Consistency and Fluidity. These metrics cover an important cross-section of human judgement, which is not represented in the automatic metrics. Specifically, we seek to quantify the naturalness and consistency of \name{} results within each dialogue span. Table~\ref{tab:human_eval} shows the performance of our models and baselines on these metrics. 

\subsection{Results Discussion}
Overall, both standalone QG models using BART and T5, as well as \name{} outperformed the SOTA baseline, CFNet, on both metrics, while the \name{} model achieved the best performance of all models on both metrics. The Answer Consistency roughly indicates that the question types were better aligned with the target answer and dialogue history, than the baseline, while the Fluency metric points to improvements in factors like grammar, coreference alignment, and dialogue flow. Together with the Automatic Metrics, these results support our finding that the \name{} model is able to learn to produce conversational dialogue that is well aligned with the CoQA dataset, both lexically and semantically, and more robust in general.

The improvement over single CQG seq2seq models like T5 demonstrates the success of learning joint question-answer representations and using the encoding of the latter to inform the QG module. We present a more complete analysis of error and ablation studies in the following sections. The answer-aware QG strategy is also validated here, as shown by the significant improvement of every answer-aware model over the answer-unaware (ReDR) baseline. Finally, the comparison of \name{} to the current answer-aware QG SOTA, CFNet, demonstrates the importance of the question-answer representation and encoding scheme in our model. While CFNet incorporates tactical model components to improve the quality of CQG along specific dimensions, such as coreference and dialogue flow, our model is able to flexibly learn the progression of questions without the need for architectural components that target specific dialogue features. Another important point is that CFNet excluded all Yes/No questions from their analysis, as they proved difficult to reliably generate. Our model not only achieves SOTA performance, but is also able to natively generate every question type present in the dataset. Moreover, we notice that our model shows improved coreference alignment ability when generating questions with complex and entangled dialog history.

\subsection{Ablation Study}

\begin{table*}
    \centering
     \begin{tabular}{c c c c c c c c c c} 
     \hline
      Model & P & B1& B2& B3& B4 & M & RL & BS & MS \\
     \hline

      \name{}-M & 7.04 & 53.15 & 35.31 & 23.31 & 15.78 &  40.15 & 50.98 & 93.14 & 36.40 \\
      
      \name{}-S & 9.55  & 49.26 & 31.06 & 20.24 & 12.11 & 33.26 & 46.23 & 92.53 & 32.82 \\
     
     \hline
     
      M w/o history & 9.13 & 45.53 & 28.35 & 18.31 & 11.27 &  30.35 & 40.45 & 92.54 & 27.59 \\
      
      S w/o history & 11.1 & 42.54 & 25.63 & 14.91 & 8.01 & 27.73 & 39.23 & 91.03 & 24.30 \\
     \hline
     
      M w/o  highlight & 7.83 & 47.07 & 30.63 & 20.54 & 12.91 &  32.56 & 43.63 & 92.36 & 31.50 \\
      
      S w/o  highlight & 11.09 & 43.63 & 25.74 & 17.14 & 10.54 & 27.73 & 40.43 & 91.23 & 24.82 \\

      \hline
      M w/o AQ order & 7.43 & 50.23 & 33.65 & 21.43 & 14.08 & 37.35 & 48.88 & 92.94 & 34.73 \\
      
      S w/o AQ order & 9.90 & 47.65 & 30.51 & 19.04 & 10.80 & 31.58 & 42.82 & 92.21 & 29.74 \\
      \hline
      
      \hline
      M w/o AE module & 8.05 & 51.64 & 33.26 & 21.26 & 13.86 & 37.23 & 47.23 & 92.64 & 32.23 \\
      
      S w/o AE module  & 15.21 & 45.23 & 28.19 & 18.01 & 10.73 & 29.43 & 44.12 & 92.13 & 27.28 \\
      \hline

    \end{tabular}
    \caption{Ablation Study Results. }
    \label{tab:ablation}
\end{table*}

Our ablation study aims to understand the effectiveness of various design choices in the \name{} approach outlined ~(in Section~\ref{sec:AEQG}). In all ablation experiments, the reference model is the \name{} model, combining the AE and QG modules. Results of this analysis are shown in Table~\ref{tab:ablation}. We have applied the following ablations:

\subsubsection{Removing the dialogue history}
Here, we evaluate the effect of the flow propagation training scheme. Table~\ref{tab:ablation} shows that removing the dialogue history, and consequently, any notion of dialogue flow, reduces performance across all the metrics~(e.g., approximately 14\% for both small and medium versions). These results match the intuition that dialogue history provides essential context to correctly handle coreferences and natural transitions.

\subsubsection{Removing the answer rationale highlight tokens}
To evaluate the effect of grounding the generated questions in the relevant passage text, we remove the answer rationale highlight tokens from the input passage. The results in Table~\ref{tab:ablation} show that this ablation decreases performance in all the metrics. For example, removing the highlight reduces BLEU-1 for the medium GPT from 53.15 to 47.07~(approximately 11\% reduction). We conclude that the highlighted tokens ground the model in the relevant passage information, providing essential context while focusing the scope of the question.

\subsubsection{Changing the order of the questions and answers}
As discussed in Section~\ref{sec:preprocess}, we have used the AQ order instead of QA in our input encoding. Here, we evaluate the effect of such ordering. As Table~\ref{tab:ablation} shows, reversing the order of the question and answers results in a performance reduction of approximately 5\% in the BLEU-1 score for the medium GPT model. This shows that the AQ order is a more natural structure for dialogue flow propagation, since answers precede the generated question in each turn.

\subsubsection{Removing the AE module} As discussed in Sections~\ref{sec:intro} and ~\ref{sec:AEQG}, by including the AE module, we aim to address the challenge of expressing the representations of questions and answers over multiple dialogue turns. Here, we remove the AE module to validate the effect of this modelling choice. The results in Table~\ref{tab:ablation} show that removing the AE module reduces the performance of the model by 3\% and 8\% in the BLEU-1 score for the medium and small GPT models, respectively. This indicates that propagating the question-answer representations across dialogue turns produces rich temporal representations that improve the fidelity of dialogue flow.

\subsection{Error Analysis}
In order to better understand the performance differentiation between our model and baselines considered here, we inspected some samples of generated questions with poor quality.
While the SOTA baseline, CFNet, neglected all Yes/No questions completely, our model is overall, very successful at generating this type, alongside others such as factoid and explanation. However, Yes/No questions can still be problematic when the answer context includes many potential targets, each of which could be satisfied by a consistent Yes/No question.
We also find that in minority cases, \name{} cannot handle questions requiring complex logic or reasoning to arrive at the target answer. We hypothesize that a more powerful pre-training model could alleviate this issue.
Also, the \name{} model sometimes includes additional details related to the answer, not contained in the gold question, which results in slightly more verbose, though consistent, questions.

\section{Discussion and Ethical Issues}\label{sec:ethical}

The results presented here demonstrate the efficacy of modern Transformer-based architectures, and specifically \name{}, in producing conversational questions on a challenging dataset. While answer-aware QG is our focus here, we plan to expand this in future work to include answer-unaware and open-ended QG, multi-task NLG involving QA, and domain-specific dialogue simulation. The flexibility of the \name{} architecture lends itself well to each of these problems, as representations from different inputs and tasks can be shared easily between modules.

As for the practical implications of our QG work, a number of applications mentioned in previous sections could immediately take advantage of QG features, either for training QA models or generating user-facing questions. In the former setting, generation models, such as \name{}, risk polluting the training dataset with examples that are noisy or inconsistent with the target answers, which can cause undesirable effects at inference time. In the latter setting, generation models may suffer from bias based on the questions available for training, which may lead to misrepresentation of application domains and individual users. Additional work is required to understand the extent of these issues in real-world applications, and identify corrective measures to ensure model robustness and diversified training distributions.

\section{Conclusion}\label{sec:conclusion}
In this paper, we introduce \name{}, an answer-aware Conversational Question Generation model that outperforms all baselines on both automatic and human evaluation metrics on the inverted CoQA dataset~(e.g., BLEU-1 improvement of 48\% and 28\% with GPT medium compared to ReDR and CFNet, respectively). We have designed a two-stage GPT-2-based architecture that jointly learns passage and dialogue history representations via a flow propagation training method. \name{} produces high-quality questions in multi-turn dialogue, addressing previous SOTA issues such as question type fidelity, question-answer inconsistency and coreference misalignment. Finally, we have performed and presented extensive ablation studies for various aspects of our approach.

\bibliography{eacl2021}
\bibliographystyle{acl_natbib}

\appendix

\end{document}